\documentclass[sigconf,nonacm]{acmart}

\setcopyright{none}
\renewcommand\footnotetextcopyrightpermission[1]{}
\usepackage{pifont}    
\usepackage{enumitem}  

\begin{document}

\title{Scrutinizing the Scrutinizer: Reward-Aligned RAG Generators Learn to Game Reference-Free Judges}

\author{Cheruvu Manikantha Sai Abhinav}
\affiliation{%
  \institution{Department of Electrical and Computer Engineering, Duke University Pratt School of Engineering}
  \city{Durham}
  \state{North Carolina}
  \country{United States}
}
\affiliation{%
  \institution{Birla Institute of Technology \& Science Pilani}
  \city{Hyderabad}
  \country{India}
}

\author{Ricardo Accorsi Casonatto}
\affiliation{%
  \institution{School of Industrial and Systems Engineering, Georgia Institute of Technology}
  \city{Atlanta}
  \state{Georgia}
  \country{United States}
}
\affiliation{%
  \institution{Department of Surgery, Department of Anesthesiology, Duke University School of Medicine}
  \city{Durham}
  \state{North Carolina}
  \country{United States}
}

\author{Rishikesan Kamaleswaran}
\affiliation{%
  \institution{Department of Electrical and Computer Engineering, Duke University Pratt School of Engineering}
  \city{Durham}
  \state{North Carolina}
  \country{United States}
}
\affiliation{%
  \institution{Department of Surgery, Department of Anesthesiology, Duke University School of Medicine}
  \city{Durham}
  \state{North Carolina}
  \country{United States}
}

\author{Aditya Nagori}
\affiliation{%
  \institution{Department of Surgery, Department of Anesthesiology, Duke University School of Medicine}
  \city{Durham}
  \state{North Carolina}
  \country{United States}
}

\begin{abstract}
Reference-free judges are increasingly used not only to evaluate
retrieval-augmented generation (RAG) systems but to guide their training. We
align RAG generators with GRPO and GDPO on inexpensive lexical proxies of
the criteria these judges score, and show that this does not reliably improve
answer quality and instead produces outputs that game the judges. We report
three findings.
First, lower score coverage: RAGAS Faithfulness returns no score for
28--41\% of answers from heavily copying policies, versus 4--7\% of base answers to the same
questions, so its averages cover a smaller share of those answers. Second, a
controlled result: in a 2$\times$2 factorial over six matched seeds, a
saturating format penalty increased verbatim context copying (copy rate 0.99,
versus 0.27 for cleaning alone), so a common anti-gaming fix can make copying
worse. Third, an audit against independent ground truth: on PubMedQA, counting
only answers a policy states itself, policies trained with the pipeline's own
trainer average 0.21 accuracy against 0.51 for base, and higher copying is
associated with lower accuracy ($r=-0.74$ across twelve retrained policies),
although a reference-free judge would rate the copied outputs as highly
faithful. Tagged question-echo leaves accuracy intact, so not all judge gaming
is equally harmful. The
copy attack recurs across three model families, a general-domain benchmark,
and the reward weightings we tested that credit answer--context overlap, and it
reproduces at 100\% on papers published after each model's training cutoff. Accuracy is lost where copying displaces the answer; where the policy states an
answer first, the answer survives and the harm is output pollution. Judge scores should be audited before they guide training.
\end{abstract}



\maketitle

\section{Introduction}
Evaluation validity is not a property of a judge in isolation; it is a property
of a judge in a system. When an LLM judge~\citep{llmjudge} or an inexpensive proxy for what it scores, supplies the training signal for
methods such as DPO and GRPO, its error profile is not merely observed
downstream. It is optimized against, and its blind spots are
compiled into model weights. This paper studies a closely related setting in
retrieval-augmented generation~\citep{rag}, where reference-free metrics such as
RAGAS~\citep{ragas}, G-Eval~\citep{geval}, and MiniCheck~\citep{minicheck} are
used both to score generators and, increasingly, to align them.

We find that aligning a RAG generator with standard methods
(GRPO~\citep{grpo} and GDPO~\citep{gdpo}) on innocuous proxy
rewards for the criteria these judges score (overlap with the retrieved context,
agreement with the reference, and brevity) does not reliably improve answer
quality; DPO~\citep{dpo}, trained on pairs of correct and incorrect answers,
is included as a comparison arm. The judges themselves are not in the training loop. Even so, training discovers
two behaviors that inflate their scores: verbatim copying of retrieved context, which inflates
faithfulness, and tagged question-echo, which inflates reverse-question relevance.
The apparent alignment gains are largely metric
inflation rather than quality. Our contributions concern two failure modes:
the judges' sensitivity to surface form over meaning, and training that exploits
it.

\begin{itemize}[leftmargin=1.2em,itemsep=1pt,topsep=2pt]
\item \textbf{C1: lower score coverage on copied answers}
(Section~\ref{sec:c1}). RAGAS Faithfulness returns no score for 28--41\% of
answers from heavily copying policies versus 4--7\% of base answers to the same questions.
Complete-case averages drop these rows, so score coverage should be reported
beside the score.
\item \textbf{C2: a format penalty that increases copying}
(Section~\ref{sec:c2}). In a 2$\times$2 factorial, a saturating tag penalty
increased verbatim copying (by $0.59$ on average within matched seeds), while
answer cleaning showed no consistent effect.
\item \textbf{C3: accuracy against independent ground truth}
(Section~\ref{sec:c3}). On an external benchmark with independent ground truth,
strict accuracy is 0.51 for base and 0.21 for policies trained with the
pipeline's own trainer, and higher copying is associated with lower accuracy,
although a reference-free judge would rate the copied outputs as highly
faithful; where an answer is stated before the copied text it survives. Tagged echo shows no
such drop, which helps isolate which judge
failures matter downstream.
\end{itemize}

Table~\ref{tab:map} maps each claim to the experiment it rests on. We further show that the copy attack generalizes across model families and to a
corpus of post-cutoff papers the models never saw (Section~\ref{sec:gen}). We
note one correction to a natural assumption: reference-based BERTScore is not a
clean copy control, since it too is inflated by copying (Section~\ref{sec:c1}).
Recent and concurrent studies~\citep{veritas,asc} report related
faithfulness-reward hacking in RAG; our primary contributions are the score-coverage result and the defense
factorial, with the quality audit (C3) characterizing \emph{when}
the attack degrades delivered output and when the buried answer still survives.

\section{Related work}
Reward hacking of learned and LLM-based evaluators is a well-documented
concern~\citep{goodhart,veritas}, and the verbatim-copy weakness of faithfulness
metrics has been noted in principle~\citep{factmetrics}. Our contribution
differs in three ways. Standard alignment training, rather than a hand-crafted
perturbation, discovers and exploits these weaknesses in a working RAG
pipeline. We identify a coverage failure, in which the judge returns no score for a large
share of attacked outputs rather than merely mis-scoring them. And we isolate,
in a factorial, which component of a common defense increases copying. Concurrent
work~\citep{asc} trains long-form RAG generators with GRPO on a checker-based
faithfulness reward, finds that they learn to copy passage sentences, and
reports that KL and copy penalties move the hack rather than remove it. Our
factorial isolates a different defense, a format penalty, and our score-coverage
and scoring analyses are complementary to that study. Prior work on
preference optimization for RAG~\citep{rpo} aims to improve answer quality and
does not study whether the resulting policies game the evaluation metrics. For
objective ground truth we use PubMedQA~\citep{pubmedqa}, a retrieval-native
benchmark that is also part of the MIRAGE suite~\citep{mirage}.

\begin{table*}[t]
\centering
\caption{Claim-to-experiment map: which result each claim rests on.}
\label{tab:map}\small
\begin{tabular}{llll}
\toprule
Claim & Eval set & Model(s) & Metric / test \\
\midrule
C1 score coverage (28--41\% vs 4--7\%) & small+scaled KG & Llama-3.1-8B & RAGAS unscored rate; McNemar \\
C2 penalty increases copying & KG60 & Llama GDPO & copy rate, $2\times2$, 6 matched seeds \\
C3 higher copying, lower accuracy & PubMedQA (500) & base + 18 trained policies & strict and judge-assisted accuracy \\
Second domain & BoolQ (500) & base/GRPO & accuracy + copy rate \\
Reward weighting & PubMedQA & GRPO variants & copy rate, output length \\
Cross-model & small KG (22 answerable) & Qwen, Mistral & copy-rate delta \\
Not memorization & 139 post-cutoff papers (46 KG) & Llama-3.1-8B & copy rate \\
\bottomrule
\end{tabular}
\end{table*}

\section{Experimental setup}
\label{sec:setup}
\textbf{Generators.} Llama-3.1-8B-Instruct (primary), Qwen2.5-7B-Instruct, and
Mistral-7B-Instruct-v0.3. \textbf{Alignment.} DPO, GRPO, and GDPO via TRL~1.1.0,
with GDPO using the library's native normalize-then-sum aggregation. GRPO
and GDPO optimize the reward below; DPO is trained on pairs of correct and
incorrect answers.
\textbf{Reward.} The proxy reward with no anti-gaming defense, which we call the gameable
reward: $0.4\cdot$ROUGE-L(answer, context) $+\,0.4\cdot$token-F1(answer,
ground truth) $+\,0.2\cdot$conciseness (ROUGE-L as in~\citep{rouge}). The post-cutoff replication in
Section~\ref{sec:gen} uses the pipeline's equivalent no-defense variant with
conciseness weight $0.1$. \textbf{Judges scored.} RAGAS 0.4.3 (Faithfulness and
ResponseRelevancy, with a 70B judge), DeepEval G-Eval (faithfulness and
relevance rubrics), MiniCheck (NLI), and BERTScore~\citep{bertscore} (reference-based).
\textbf{Pipeline, corpora and evaluation sets.} The RAG system is the
open-source agentic hybrid framework of our earlier work~\citep{hybridrag}. A small (11-paper) and a scaled (roughly
150-paper) biomedical corpus, each split into knowledge-graph (KG, Cypher) and
vector-store (VS, similarity) questions. Evaluation uses document-disjoint
holdouts: 220 questions on the small corpus (144 VS, 76 KG) and 136 on the scaled
corpus (72 VS, 64 KG), a separate 60-question KG holdout (KG60) for the defense
factorial and the small-corpus KG comparisons of Section~\ref{sec:c1}, 500 PubMedQA items, and a 64-question KG holdout on the post-cutoff
corpus. Each
answer is scored once per question. \textbf{Copy classifier.} A single classifier, frozen
before the reported runs were scored, labels an answer a substantive copy when more than
90\% of its alphanumeric tokens appear in the retrieved context, excluding empty
sentinels and short extractions. It measures token overlap only: a short answer assembled from the
context's own words also counts, which is why the base model has a non-zero
rate ($0.34$ on PubMedQA). \textbf{Evaluation hygiene.} Native-corpus
evaluations use greedy decoding with a repetition penalty matched between
fine-tuned checkpoints and the served baseline, and benchmark evaluations use
plain greedy decoding for every model; prompts are tokenized without re-adding
special tokens (avoiding a family-dependent double-BOS asymmetry); metric
scoring is checkpointed and resumable, so transient judge errors are retried
rather than frozen into results; and model snapshots are hash-verified against
upstream weights before training. \textbf{Infrastructure.} Training and
evaluation run on NVIDIA H200 GPUs under SLURM; the generators are trained and
run through HuggingFace TRL and \texttt{transformers}, while the 70B evaluation
judge (Llama-3.3-70B-Instruct) and the embedding model (Qwen3-Embedding-8B) are
served locally through Ollama.

\section{C1: Lower score coverage on copied answers}
\label{sec:c1}
We score stored answers (base versus aligned) with each metric, on both the raw
answer and a cleaned answer with echo and scaffolding stripped, and test two
mechanisms: context copying and tagged echo. Restating the question in prose
also raises relevance, but more for base than for the aligned policies, so we do
not treat it as a learned attack. The headline finding
concerns coverage, not score. \textbf{RAGAS Faithfulness returns no score for
28--41\% of answers from the three heavily copying arms, against 4--7\% of base
answers to the same KG questions} (Table~\ref{tab:cov}). The failures are the
judge's own structured-output parses returning no usable score. Paired by
question, the difference is significant for the three heavily copying arms
(exact McNemar $p=1\times10^{-4}$ to $7\times10^{-3}$) and not for the milder
DPO arm on the scaled corpus ($p=0.11$) or for GDPO, whose unscored rate matches base. Failed rows drop out of complete-case means, so the reported
average covers a smaller share of the copying arms' answers. We limit the
conclusion to that: score coverage is lower on copied answers.

\begin{table}[t]
\centering
\caption{RAGAS Faithfulness coverage on KG questions: answers that received no
score, out of those attempted, for each aligned arm and for the base model on
the same questions.}
\label{tab:cov}
\begin{tabular}{lcc}
\toprule
Arm (corpus) & Aligned arm & Base, same questions \\
\midrule
GRPO (scaled) & 20/49 (41\%) & 2/49 (4\%) \\
GRPO (small) & 20/57 (35\%) & 4/57 (7\%) \\
DPO (small) & 15/54 (28\%) & 4/54 (7\%) \\
DPO (scaled) & 8/54 (15\%) & 2/54 (4\%) \\
GDPO (small) & 2/60 (3\%) & 4/60 (7\%) \\
GDPO (scaled) & 3/64 (5\%) & 3/64 (5\%) \\
\bottomrule
\end{tabular}
\end{table}

Several metrics are, by contrast, directly fooled by copying
(Table~\ref{tab:e1}). MiniCheck shows a $+0.33$ inflation for small-corpus GRPO
(Holm $p=4\times10^{-4}$), and G-Eval faithfulness behaves similarly. RAGAS
ResponseRelevancy is fooled by echo on the scaled corpus ($+0.058$ raw versus
cleaned, Holm-significant). Correcting an assumption we ourselves initially
held, reference-based BERTScore is also inflated by copying ($+0.03$ to
$+0.07$, all Holm-significant), because copying the retrieved context raises
lexical similarity to the short reference. No faithfulness or reference-based metric in our panel is
a clean control for copying. These metric effects also survive a global multiple-comparison correction across all $62$ panel comparisons (Benjamini--Hochberg and Holm), not only within families. The practical recommendation is to report score
coverage, that is, the fraction of outputs a judge returns a score for per
system, and not only the scores.

\begin{table*}[t]
\centering
\caption{Metric outcomes by attack mechanism (Holm-adjusted within comparison
families). Three metrics are inflated by copying and one loses coverage.}
\label{tab:e1}
\small
\begin{tabular}{lll}
\toprule
Metric & Copying $\to$ faithfulness & Echo $\to$ relevance \\
\midrule
MiniCheck (NLI) & fooled (small corpus, $+0.33$) & -- \\
G-Eval faithfulness & fooled (small corpus, $+0.22$) & -- \\
RAGAS Faithfulness & \textbf{lower coverage} (28--41\% unscored) & -- \\
RAGAS ResponseRelevancy & -- & fooled (scaled corpus, $+0.058$) \\
G-Eval relevance & -- & echo-insensitive; residual inflation \\
BERTScore (reference-based) & inflated ($+0.03$ to $+0.07$) & not significant \\
\bottomrule
\end{tabular}
\end{table*}

\section{C2: A format penalty increases copying}
\label{sec:c2}
A common anti-gaming defense combines two components: cleaning the answer
before scoring, and a saturating format penalty on echo tags. We ask how each
component affects copying. We train four matched GDPO arms on
KG questions that differ in exactly one factor, using the same base model,
training split, contexts, and six seeds (42--47) per arm, and classify copying
with the frozen classifier. Table~\ref{tab:e4} reports the substantive copy
rate.

\begin{table}[t]
\centering
\caption{Anti-gaming factorial (GDPO, KG questions). The tag penalty alone, and
not the cleaning, reproduces the full defense's copy rate.}
\label{tab:e4}
\begin{tabular}{lccc}
\toprule
Arm & Clean? & Penalty? & Copy rate \\
\midrule
no defense & \ding{55} & \ding{55} & 0.543 (seed-noisy) \\
clean-only & \ding{51} & \ding{55} & 0.266 \\
penalty-only & \ding{55} & \ding{51} & \textbf{0.994} \\
both (full defense) & \ding{51} & \ding{51} & 1.000 \\
\bottomrule
\end{tabular}
\end{table}

Penalty-only (0.994) matches the full defense (1.000), while cleaning-only
(0.266) does not exceed the no-defense arm (0.543, which varies widely across
seeds). The tag penalty increased copying. Matched by seed, both penalty arms
copy at least as much as both no-penalty arms in every one of the six seeds;
the mean within-seed difference between penalty-on and penalty-off is $0.59$
(Wilcoxon signed-rank $p=0.031$, the smallest value six pairs allow), and
cleaning shows no consistent effect ($p=0.44$). Across seeds the penalty arms
copy at 0.97 to 1.00 and the no-penalty arms at 0.00 to 0.97. A plausible
mechanism is that the reward still credits answer--context overlap, and on KG
questions the retrieved context is essentially the answer, so a policy
penalized for echo tags collects the reward by copying. We did not measure echo
rates in these arms, so we do not claim that copying replaced echo. The metric
profile of the penalty arms is consistent with copying: faithfulness of 0.98
and relevance of 0.27. For practitioners, a surface penalty added to a reward
that still credits overlap can increase copying. We show this for GDPO on KG
questions, the setting where context and answer coincide; whether it holds
where they diverge, or with a different optimizer, is future work
(Section~\ref{sec:lim}).

\section{C3: Accuracy against independent ground truth}
\label{sec:c3}
Our thesis is that the judges are fooled, so the judges cannot adjudicate
answer quality. Reference-anchored metrics on our native corpus are also
confounded: on KG questions the context approximately equals the ground truth,
so copying inflates even F1 against the reference. We therefore anchor quality
on an external, human-curated benchmark with independent ground truth:
PubMedQA~\citep{pubmedqa}, a dataset also used in the MIRAGE
suite~\citep{mirage}; we use its original form, in which each question carries a gold abstract as context
and a yes/no/maybe label ($n{=}500$; base uses greedy decoding, with a
multi-seed sampled base reported below). We report two sets of runs. The
\emph{original} set is three seeds each of a GRPO and a GDPO policy, trained
with the gameable reward of Section~\ref{sec:setup}. We then
ran a \emph{replication}: twelve freshly trained policies, three seeds for each
of four training configurations.
Table~\ref{tab:expb} reports both accuracy scores and the substantive copy rate.

\begin{table}[t]
\centering
\caption{Accuracy and copy rate on PubMedQA (means over three training seeds;
base is greedy decoding). Strict accuracy counts only a label the output states
itself; judge-assisted accuracy also accepts a label inferred by the scorer.
Strict accuracy is reported for base and the replication. The pipeline trainer uses
conciseness weight $0.1$; the standalone implementation uses
weight $0.2$.}
\label{tab:expb}
\resizebox{\columnwidth}{!}{%
\begin{tabular}{lccc}
\toprule
Configuration & Strict & Judge-assisted & Copy rate \\
\midrule
base & 0.506 & 0.632 & 0.34 \\
\midrule
\multicolumn{4}{l}{\emph{Original runs}} \\
GRPO & -- & 0.466 & 1.00 \\
GDPO & -- & 0.689 & 0.21--0.52 \\
\midrule
\multicolumn{4}{l}{\emph{Replication}} \\
GRPO, pipeline trainer, no defense & 0.203 & 0.547 & 0.87 \\
GRPO, pipeline trainer, full defense & 0.215 & 0.567 & 0.82 \\
GRPO, standalone reward & 0.557 & 0.687 & 0.66 \\
GDPO, standalone reward & 0.331 & 0.593 & 0.65 \\
\bottomrule
\end{tabular}}
\end{table}

\textbf{Two scores.} We score each output two ways. \emph{Strict} accuracy
credits an output only for a label it states itself (a leading yes, no or
maybe, or one inside an answer tag); a copied output that states no label
counts as unanswered, and any other output keeps the judge-assisted verdict.
\emph{Judge-assisted} accuracy is more lenient: failing a leading label, it
accepts a label word if exactly one occurs anywhere in the output, and failing
that, a 70B judge's reading of what the output concludes. For copied context
with no label of its own, that reading comes from the abstract, not from the
policy, and is correct only $0.45$ to $0.52$ of the time. The strict rule was
defined post hoc and is applied identically to base
(Figure~\ref{fig:examples}).

\textbf{Higher copying is associated with lower accuracy.} Under the strict
score, base reaches $0.506$, while the six policies trained with the pipeline's
own trainer average $0.209$ ($0.000$ to $0.372$) and leave $72\%$ of questions
unanswered. Across the twelve replication policies, copy rate and strict
accuracy correlate at $r=-0.74$ ($p=0.006$), and policies with copy rates below
$0.6$ average $0.522$, level with base. The judge-assisted score shows the same
association ($r=-0.74$, $p=0.006$; Spearman $-0.67$), strongly within the
pipeline trainer ($r=-0.96$, $n=6$) and weakly within the standalone reward
($r=-0.56$, $n=6$, not significant). Pooled over all outputs, copied answers are
judged correct $56.1\%$ of the time against $71.0\%$ for non-copied answers
(lower in nine of the ten policies that produce both; base shows a smaller gap,
$59.2\%$ against $65.7\%$). Three of the four policies that copy at least $90\%$
of the time have judge-assisted scores of $0.464$, $0.466$ and $0.504$, at or
below the majority-class baseline of $0.55$ (the fraction of ``yes'' labels).
The fourth states an answer before the copied passage and scores $0.658$:
accuracy is lost where copying displaces the answer, not where it follows one.
The original runs fit the pattern: the GRPO policy copied the context in
essentially every case and its judge-assisted accuracy was 0.466
($0.454/0.468/0.476$) against 0.632 for base; for the seed-42 run the Wilson
95\% interval, $[0.411, 0.498]$, does not overlap the base interval
$[0.589, 0.673]$. These are associations across trained policies, not a
controlled manipulation of copying. A reference-free judge would rate the
copied outputs as highly faithful, yet most of them do not answer the question.

To confirm this is not merely an artifact of
transferring a biomedical-corpus policy to PubMedQA, we also trained the copy
attack \emph{in-domain} on a disjoint PubMedQA split (one training seed): it still copies heavily
($78\%$) and judge-assisted accuracy drops from $0.632$ to $0.518$. A multi-seed sampled base (temperature $0.7$, seeds
42/43/44) gives $0.632/0.628/0.660$ (mean $0.640$), so the gap is not an artifact
of comparing trained seeds against a single deterministic base run. The
copy rates are also insensitive to the classifier's $90\%$ overlap threshold: on
the original runs the GRPO rate stays at $0.988$--$1.000$ and the base rate at
$0.30$--$0.42$ across $80$--$95\%$ thresholds.

\begin{figure}[t]
\centering
\fbox{\begin{minipage}{0.96\columnwidth}\small
\textbf{Question} (PubMedQA 21645374; gold label \emph{yes}): \textit{Do
mitochondria play a role in remodelling lace plant leaves during programmed
cell death?}

\medskip
\textbf{Copy without an answer} (GRPO, pipeline trainer, full defense, seed 42):
``Programmed cell death (PCD) is the regulated death of cells within an
organism. The lace plant (\emph{Aponogeton madagascariensis}) produces
perforations in its leaves through PCD. [\dots]'' (the abstract reproduced
verbatim, 174 words; scored incorrect)

\medskip
\textbf{Answer, then copy} (GRPO, standalone reward, seed 43):
``\texttt{<answer>} Yes \texttt{</answer>} \texttt{<context>} Programmed cell
death (PCD) is the regulated death of cells [\dots]'' (scored correct)

\medskip
\textbf{Tagged echo} (GRPO, standalone reward, seed 42):
``\texttt{<question>} Do mitochondria play a role in remodelling lace plant
leaves during programmed cell death? \texttt{</question>}
\texttt{<context>} The following paper elucidates the role of mitochondrial
dynamics [\dots]'' (states no label; the judge-assisted scorer reads the
echoed question and context and returns \emph{yes}, scored correct; unanswered
under the strict rule)
\end{minipage}}
\caption{Abridged verbatim outputs from the replication for one question. Accuracy
is lost where copying displaces the answer, not where an answer precedes it;
the third output shows why the judge-assisted score is generous.}
\Description{A boxed text example showing one PubMedQA question and three abridged model outputs: a verbatim copy with no answer, an answer followed by a copy, and a tagged echo of the question.}
\label{fig:examples}
\end{figure}

\textbf{Tagged echo does not lower accuracy.} Policies that game the relevance metric by
wrapping the question or answer in scaffolding tags, without heavy copying, keep
their accuracy. The original GDPO runs score 0.689 judge-assisted, and the four replication
policies with copy rates below $0.6$, all of which emit such tags, average
$0.522$ under the strict score against $0.506$ for base ($0.678$ against
$0.632$ judge-assisted). We do not attribute the small gain over base to
gaming; fine-tuning on a biomedical corpus plausibly confers domain adaptation,
and our claim is only the absence of degradation. Which channel a policy falls
into is not fixed by the optimizer: in the replication it varies with the seed
and the training implementation, and one GDPO seed moved to heavy copying
($0.92$) and scored $0.466$ judge-assisted. Not all judge gaming is equally harmful, and
identifying which judge failures propagate into worse behavior, as opposed to
mere metric inflation, deserves attention in its own right. We therefore scope
the quality-degradation claim to copying.

\textbf{A general-domain benchmark: copying generalizes, and an answer stated
first survives.} To separate the copy \emph{behavior} from its \emph{quality
cost}, we trained GRPO with the gameable reward and evaluated it in-domain on
BoolQ~\citep{boolq}
(general-domain yes/no; three training seeds). The behavior generalizes: the
policy copies the passage in $98\%$ of outputs, against $4\%$ for base. Accuracy
is preserved ($0.812$ for base; $0.838/0.854/0.828$ trained), because every
output begins with a yes or no before the copied passage, the same answer-first
pattern that preserved accuracy on PubMedQA; the two scores coincide here. A reference-free judge would rate
these padded outputs as highly faithful.

\section{Generality}
\label{sec:gen}
\textbf{Across model families.} With the same gameable reward, the copy attack
recurs on Qwen2.5-7B and Mistral-7B: the KG copy rate rises from a near-zero
base floor by $+0.53$ to $+0.83$ across GRPO and GDPO arms (six seeds each). A
copy here is the raw graph result emitted verbatim in place of a sentence. The rate is computed on the 22 of 76 KG questions that return a graph result
on this 11-paper corpus. The attack is not a quirk of a single model.

\textbf{Sensitivity to the reward weighting.} Copying is not an artifact
of the specific $0.4/0.4/0.2$ reward. Training GRPO under alternative weightings
of the same components shows the attack tracks the ROUGE-L(answer, context)
overlap term (Table~\ref{tab:altreward}; means over three training seeds): a
ROUGE-L-only reward copies at $0.999$, the default weighting at $0.655$, and a balanced $0.5/0.5$ ROUGE-L/F1 reward
at $0.672$, all far above the base rate of $0.34$, whereas an
F1-against-ground-truth-only reward (which does not credit context overlap)
copies at $0.357$, close to base. Output length shows what the
rate means in each case: with overlap alone the policy dumps the passage (a
median of 167 words), and the default weighting also yields long outputs, but
under the balanced reward the outputs stay short (22 words) and are answers
phrased in the abstract's words, not dumps. The overlap term therefore raises
reliance on the context's wording in every case, and full verbatim dumping
appears when it dominates.

\begin{table}[t]
\centering
\caption{Copy rate and median output length on PubMedQA under alternative GRPO reward weightings (three training seeds, length pooled over their outputs; base $0.34$, 13 words). Copying tracks the answer--context overlap (ROUGE-L) term.}
\label{tab:altreward}
\small
\begin{tabular}{lcc}
\toprule
Reward & Copy rate & Words \\
\midrule
ROUGE-L only (overlap) & 0.999 & 167 \\
$0.4$\,ROUGE-L$+0.4$\,F1$+0.2$\,conc.\ (default) & 0.655 & 147 \\
$0.5$ ROUGE-L $+\,0.5$ F1 & 0.672 & 22 \\
F1 vs.\ ground truth only & 0.357 & 21 \\
\bottomrule
\end{tabular}
\end{table}

\textbf{Memorization ruled out.} We rebuilt the RAG corpus from 139 papers
published in 2025--26 (a corpus distinct from the scaled corpus of
Section~\ref{sec:setup}), after the release of every model we use (Llama-3.1,
documented cutoff December 2023; Qwen2.5, released September 2024;
Mistral-7B-v0.3, released May 2024), and therefore after every model's training
cutoff, so the models cannot have seen them in pretraining. On this corpus the base model copies 0\% of KG
contexts (0 of 46 questions with a graph result), while GRPO and GDPO both copy
100\% at all three seeds (30 of 30 in each run, after the classifier sets aside
16 short extractions), with
faithfulness rising to about 0.98 and VS relevance falling from 0.71 to between
0.54 and 0.69. The attack is induced by the reward, not recalled from pretraining.

\section{Toward judge deployment disclosure}
\label{sec:disc}
Reporting standards for deployed models, such as model cards~\citep{modelcards},
document intended use and known limitations, but do not yet address the case
where the model is an evaluator embedded in a training loop. Our results suggest
two concrete, inexpensive disclosure instruments for such judges. First, a \emph{score-coverage audit}: report the
fraction of outputs the judge fails to score, broken down by the system under
evaluation, since coverage can differ sharply between systems
(Section~\ref{sec:c1}) and complete-case averaging otherwise hides it. Second, a \emph{frozen artifact classifier}: fix and publish the rule that
labels gamed outputs (for example, a copy classifier like the one in
Section~\ref{sec:setup})
before outcomes are inspected, so behavioral rates cannot be tuned post hoc.
Both are pipeline-agnostic, and neither requires additional model calls.

\section{Limitations and future work}
\label{sec:lim}
Quality here is anchored on objective benchmark accuracy; human ratings of
open-ended answers on the native corpus are future work. The accuracy audit uses
one retrieval-native benchmark, PubMedQA, with BoolQ as a second domain.
Extending it to closed-book medical QA (MedQA, MMLU-Med, MedMCQA) needs a
neutral prompt: under the pipeline's retrieval prompt those items draw refusals
at rates that differ across configurations, so we draw no conclusions from our
runs on them. The study centers on one RAG pipeline; the external benchmarks and
the post-cutoff corpus widen its scope. The knowledge-graph results use the
questions that return a graph result (22 on the small corpus, 46 on the
post-cutoff corpus), and larger graphs would tighten them. The original PubMedQA result is accompanied by a replication with freshly trained policies, which
shows more variation across seeds. Several extensions remain. The defense factorial should be repeated with GRPO and
on vector-store questions; the one comparison we have, GRPO on PubMedQA
(Table~\ref{tab:expb}), shows similar copy rates with and without the defense.
Score coverage should be mapped across question types, since on vector-store
questions of the scaled corpus the metric also fails on 28\% of base answers.
Finally, the rewards studied here are lexical proxies. Whether rewards computed
by a judge behave the same way is open: in one preliminary run with an 8B-judge
faithfulness reward we saw almost no copying (2\%), and a policy trained
directly on a proxy of the relevance metric raised that metric by $0.07$ on the
scaled holdout.

\section{Conclusion}
Aligning a RAG generator on proxy rewards for what reference-free judges measure
does not reliably yield better answers; the score gains the judges then report
are largely metric inflation. Training teaches the generator to game the judge,
and a natural defense can increase copying. Judges embedded in alignment pipelines should
be audited for score coverage and for what happens downstream when they fail, not
only for score accuracy. These failures are visible only when the judge is
treated as part of the training system rather than in isolation.

\section{Safe and responsible innovation statement}
This study uses only publicly available scientific papers and public benchmarks
(PubMedQA and BoolQ) and no personal or patient data, and all models were run
on institutional compute without external APIs. The behaviors we document are ways a generator can inflate automatic
evaluation scores. We report them so that practitioners can audit judges before
relying on them, and the two disclosure instruments of
Section~\ref{sec:disc} are intended as low-cost safeguards. These failure modes
arise from standard training on ordinary rewards and need no specialized
technique, so we judge the benefit of disclosure to outweigh the risk of misuse.

\section*{Acknowledgments}
We thank the KamaleswaranLab at Duke University for support, and acknowledge
compute provided by the NCShare cluster (NVIDIA H200 GPUs).

\end{document}